\documentclass[10pt,twocolumn,letterpaper]{article}
\pdfoutput=1
\usepackage{times}
\usepackage{graphicx}
\usepackage{amsmath}
\usepackage{amssymb}
\usepackage{xspace}
\usepackage{xcolor}
\usepackage{enumitem}
\usepackage{listings}
\usepackage{tabularx}
\usepackage{multirow}
\usepackage[export]{adjustbox}    
\usepackage{tikz}

\usepackage{comment}

\usepackage{subfig}

\usepackage{animate}

\usepackage[numbers,square,sort&compress]{natbib}
\usepackage[
pagebackref=true,
breaklinks=true,
letterpaper=true,
colorlinks,
bookmarks=false]{hyperref}

\newcommand{\figref}[1]{Fig.~\ref{#1}}

\newcommand{\sref}[1]{Sect.~\ref{#1}}

\newcommand{\ve}[1]{\mathbf{#1}} 

\newcommand{\real}{\mathbb{R}}
\newcommand{\bx}{\mathbf{x}}

\newcommand{\argmax}{\operatornamewithlimits{argmax}}

\usepackage{xfrac}
\usepackage{color}

\newcommand{\blockSpeedVisCap}[5]{
	\subfloat[{\tiny  appearance $#5=0$}]{
	\begin{tikzpicture}%
		\node  {\rotatebox{90}{\tiny \hspace{-0.00\textwidth} #1 #3}};
	\end{tikzpicture}%
		\includegraphics[width=0.09\textwidth]{figures/#2/#3_rgb-0}}
	\subfloat[{\tiny flow  $#5=10$}]{\animategraphics[loop,autopause,width=0.09\textwidth]{\theFPS}{figures/#2/#3-flow_mag_#4-}{0}{9}}
	\subfloat[{\tiny flow  $#5=5$}]{\animategraphics[loop,autopause,width=0.09\textwidth]{\theFPS}{figures/#2/#3-flow_mag_#4x05-}{0}{9}}
	\subfloat[{\tiny flow  $#5=1$}]{\animategraphics[loop,autopause,width=0.09\textwidth]{\theFPS}{figures/#2/#3-flow_mag_#4x01-}{0}{9}}	
	\subfloat[{\tiny flow  $#5=0$}]{\animategraphics[loop,autopause,width=0.09\textwidth]{\theFPS}{figures/#2/#3-flow_mag-}{0}{9}}	
	
}
\newcommand{\blockSpeedVisNoCap}[4]{
	\vspace{-20pt}	
	\subfloat[]{
		\begin{tikzpicture}%
		\node  {\rotatebox{90}{\tiny \hspace{-0.00\textwidth} #1 #3}};
		\end{tikzpicture}%
		\includegraphics[width=0.09\textwidth]{figures/#2/#3_rgb-0}}
	\subfloat[]{\animategraphics[loop,autopause,width=0.09\textwidth]{\theFPS}{figures/#2/#3-flow_mag_#4-}{0}{9}}
	\subfloat[]{\animategraphics[loop,autopause,width=0.09\textwidth]{\theFPS}{figures/#2/#3-flow_mag_#4x05-}{0}{9}}
	\subfloat[]{\animategraphics[loop,autopause,width=0.09\textwidth]{\theFPS}{figures/#2/#3-flow_mag_#4x01-}{0}{9}}	
	\subfloat[]{\animategraphics[loop,autopause,width=0.09\textwidth]{\theFPS}{figures/#2/#3-flow_mag-}{0}{9}}	
	
}

\newcommand{\blockSpeedVisCapUniform}[5]{
	\subfloat[{\tiny  appearance $#5=10$}]{
		\begin{tikzpicture}%
		\node  {\rotatebox{90}{\tiny \hspace{-0.00\textwidth} #1 #3}};
		\end{tikzpicture}%
		\includegraphics[width=0.09\textwidth]{figures/#2/#3_rgb_#4}}
	\subfloat[{\tiny flow  $#5=10$}]{\animategraphics[loop,autopause,width=0.09\textwidth]{\theFPS}{figures/#2/#3-flow_mag_#4-}{0}{9}}
	\subfloat[{\tiny flow  $#5=5$}]{\animategraphics[loop,autopause,width=0.09\textwidth]{\theFPS}{figures/#2/#3-flow_mag_#4x05-}{0}{9}}
	\subfloat[{\tiny flow  $#5=2.5$}]{\animategraphics[loop,autopause,width=0.09\textwidth]{\theFPS}{figures/#2/#3-flow_mag_#4x025-}{0}{9}}	
	\subfloat[{\tiny flow  $#5=1$}]{\animategraphics[loop,autopause,width=0.09\textwidth]{\theFPS}{figures/#2/#3-flow_mag_#4x01-}{0}{9}}	
	
}

\newlength\Hoffset
\newlength\Voffset
\usepackage{cvpr}
 \cvprfinalcopy 

\def\cvprPaperID{***} 

\author{
		\hspace*{-12pt}Christoph Feichtenhofer\thanks{C. Feichtenhofer made the primary contribution to this work and therefore is listed first. Others contributed equally and are listed alphabetically.}\\
		\hspace*{-12pt}TU Graz\\
		\hspace*{-12pt}{\tt \small \href{mailto:feichtenhofer@tugraz.at}{\textcolor{black}{feichtenhofer@tugraz.at}}} 
	\and	
		\hspace*{-10pt}Axel Pinz\\
		\hspace*{-10pt}TU Graz\\
		\hspace*{-10pt}{\tt \small \href{mailto:axel.pinz@tugraz.at}{\textcolor{black}{axel.pinz@tugraz.at}}}
	\and 
		\hspace*{-7pt}Richard P. Wildes\\
		\hspace*{-7pt}York University, Toronto\\
		\hspace*{-7pt}{\tt \small  \href{mailto:wildes@cse.yorku.ca}{\textcolor{black}{wildes@cse.yorku.ca}}}
	 \and 
		 \hspace*{-2pt}Andrew Zisserman \hspace*{-15pt}\\
		 \hspace*{-2pt}University of Oxford \hspace*{-15pt}\\
		 \hspace*{-2pt}{\tt \small  \href{mailto:az@robots.ox.ac.uk}{\textcolor{black}{az@robots.ox.ac.uk}}} \hspace*{-15pt}
}

\newcounter{FPS}
\begin{document}
	\title{What have we learned from deep representations for action recognition?}
	\maketitle	
	\begin{abstract}

As the success of deep models has led to their deployment in all areas
of computer vision, it is increasingly  important  to understand how these
representations work and what they are capturing.  In this paper, we
shed light on deep spatiotemporal representations by visualizing what
two-stream models have learned in order to recognize actions in video. We
show that local detectors for appearance and motion objects arise to
form distributed representations for recognizing human actions.  Key
observations include the following. First, cross-stream fusion enables
the learning of true spatiotemporal features rather than simply
separate appearance and motion features. Second, the networks can learn local representations that
are highly class specific, but also generic representations that can
serve a range of classes.  Third, throughout the
hierarchy of the network, features become more abstract and show
increasing invariance to aspects of the data that are unimportant to
desired distinctions (\eg motion patterns across various speeds). Fourth, visualizations can be used not only to shed light on learned
representations, but also to reveal idiosyncracies of training data
and to explain failure cases of the system.  \textbf{This document is best viewed offline where figures play on click.}

	\end{abstract}

	\section{Motivation} Principled understanding of how deep networks
operate and achieve their strong performance significantly lags behind
their realizations. Since these models are being deployed to all
fields from medicine to transportation, this issue becomes of ever
greater importance.  Previous work has yielded great advances in
effective architectures for recognizing actions in video, with
especially significant strides towards higher accuracies made by deep
spatiotemporal models
\cite{C3DICCV2015,feichtenhofer2016convolutional,Simonyan14b,WangECCV16,carreira2017quo}. However,
what these deep networks actually learn remains unclear, since their
compositional structure makes it difficult to reason explicitly about
their learned representations. In this paper we propose
spatiotemporally regularized activation maximization to visualize deep
two-stream representations \cite{Simonyan14b} and better understand
what the underlying models are capturing.

\begin{figure}[t]
    
    \captionsetup[subfigure]{position=bottom}
    \captionsetup[subfloat]{captionskip=0pt}
    \centering 
   \subfloat[]{\includegraphics[height=0.1\textwidth]{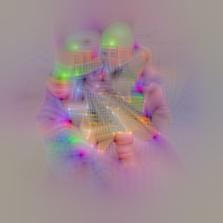}}    
    \subfloat[]{\animategraphics[loop,autoplay,height=0.1\textwidth]{\theFPS}{figures/conv5_fusion/billiard/slow/f021-flow_mag-}{0}{9}}\hfill
  \subfloat[]{\animategraphics[loop,autoplay,poster=17,height=0.1\textwidth]{\theFPS}{figures/conv5_fusion/billiard/clip/frame000}{105}{130}\label{fig_bil_rgb_ref}}
  \subfloat[]{\animategraphics[loop,autoplay,poster=17,height=0.1\textwidth]{\theFPS}{figures/conv5_fusion/billiard/mag/frame000}{105}{130}\label{fig_bil_mag_ref}}\\

    \caption{Studying a single filter at layer conv5\_fusion: (a) and (b) show what
maximizes the unit at the input: multiple coloured blobs in the
appearance input (a) and moving circular objects at the motion input
(b).  (c) shows a sample clip from the test set,  and (d) the corresponding
optical flow 
(where the
RGB channels correspond to the horizontal, vertical and magnitude flow
components respectively). Note that (a) and (b) are optimized from white noise
under regularized spatiotemporal variation.
\color{cyan}{\bf  Best viewed in Adobe Reader where (b)-(d) should play as videos.}
    }
    \label{fig_vis_billiard}
    \vspace{-15pt}
\end{figure}

As an example, in \figref{fig_vis_billiard} we highlight a single interesting unit at
the last convolutional layer of the VGG-16 Two-Stream Fusion model
\cite{feichtenhofer2016convolutional}, which fuses appearance and
motion features. We visualize the appearance and motion inputs
that highly activate  this  filter. When looking at the inputs, we observe that this
filter is activated by differently coloured blobs in the appearance
input and by linear motion of circular regions in the motion
input. Thus,  this unit could support recognition of the
Billiards class in UCF101,  and we show in \figref{fig_bil_rgb_ref} a
sample Billiards clip from the test set of UCF101. Similar to
emergence of object detectors for static 
images~\cite{zhou2014object,netdissect2017}, here we see the  emergence of a
spatiotemporal representation for an action. While \cite{zhou2014object,netdissect2017}
automatically assigned concept labels to learned internal
representations by reference to a large collection of labelled input
samples, our work instead is concerned with visualizing the
network's internal representations without appeal to any signal at the
input and thereby avoids biasing the visualization via appeal to a
particular set of samples.

Generally, we can understand deep networks from two viewpoints. First,
the \textit{architectural viewpoint} that considers a network as a
computational structure (\eg a directed acyclic graph) of mathematical
operations in feature space (\eg affine scaling and shifting, local
convolution and pooling, nonlinear activation functions, etc.).  In
previous work, architectures (such as Inception \cite{szegedy2014going}, VGG16 \cite{Simonyan15}, ResNet \cite{He16}) have
been designed by composing such computational structures with a
principle in mind (e.g.\ a direct path for backpropagation in ResNet).
We can thus reason about their expected predictions for given input
and the quantitative performance for a given task justifies their
design, but this does not explain how a network actually arrives at
these results. The second way to understand deep networks is the
\textit{representational viewpoint} that is concerned with the learned
representation embodied in the network parameters. Understanding these
representations is inherently hard as recent networks consist of a
large number of parameters with a vast space of possible functions
they can model. The hierarchical nature in which these parameters are
arranged makes the task of understanding complicated, especially for
ever deeper representations.  Due to their compositional structure it
is difficult to explicitly reason about what these powerful models
actually have learned.

In this paper we shed light on deep spatiotemporal networks by
visualizing what excites the learned models using activation
maximization by backpropagating on the input.  We are the first to
visualize the hierarchical features learned by a deep motion
network. Our visual explanations are highly intuitive and provide
qualitative support for the benefits of  separating  into two pathways when processing
spatiotemporal information -- a principle that has also been found in
nature where numerous studies suggest a corresponding separation into
ventral and dorsal pathways of the brain \cite{mishkin1983object,
Goodale92,felleman1991distributed} as well as the existence of
cross-pathway connections
\cite{saleem2000connections,kourtzi2000activation}.

\section{Related work on visualization}
\label{sec_vis_related_work}
\label{sec:related_work}

The current  approaches to visualization can be grouped into three types, 
and we review each of them in turn.

\noindent \textbf{Visualization for given inputs} 
have been used in several approaches to increase the understanding of
deep networks. A straightforward approach is to record the network
activities and sample over a large set of input images for finding the
ones that maximize the unit of interest
\cite{Zeiler13,zhou2014object,zhou2016learning,netdissect2017}. Another
strategy is to use backpropagation to highlight salient regions of the
hidden units
\cite{Simonyan14a,mahendran2016salient,zhang2016top,Selvaraju2016}. Our
method is more closely related to inspection approaches without given input.

\noindent \textbf{Activation maximization} has been used by backpropagating on, and applying gradient ascent to, the input to find an image that increases the activity of some neuron of interest \cite{Erhan09}. The method was employed to visualize units of Deep Belief Networks (DBNs)~\cite{Hinton06,Erhan09} and adopted for deep auto-encoder visualizations in \cite{Le12}. The activation maximization idea was first applied to visualizing ConvNet representations trained on ImageNet \cite{Simonyan14a}. That work also showed that the
activation maximization techniques generalize the deconvolutional
network reconstruction procedure introduced earlier \cite{Zeiler13}, which can be viewed as a special case of one iteration in the gradient based activation maximization. In an unconstrained setting, these methods can exploit the full dimensionality of the input space; therefore, plain gradient based optimization on the input can generate images that do not reflect natural signals. Regularization techniques can be used to compensate for this deficit. In the literature, the following regularizers have been applied to the inputs to make them perceptually more interpretable: $L2$ norms \cite{Simonyan14a}, total-variation norms \cite{Mahendran16}, Gaussian blurring, and suppressing of low values and gradients \cite{yosinski2015understanding}, as well as spatial sifting (jittering) of the input during optimization,  \cite{mordvintsev2015inceptionism}. Backpropagation on the input has also been used to find salient regions for a given input \cite{springenberg2014striving,zhang2016top,mahendran2016salient}, or to ``fool'' networks by applying a perturbation to the input that is hardly perceptible to humans \cite{Szegedy14,nguyen2015deep}. 

\noindent  \textbf{Generative Adversarial Networks (GANs)} \cite{goodfellow2014generative} provide even stronger natural image priors, for visualizing class level representations \cite{nguyen2016synthesizing,nguyen2016plug} in the activation maximization framework.
These methods optimize a high-dimensional code vector (typically fc\_6 in AlexNet) that serves as an input to the generator which is trained with a perceptual loss \cite{dosovitskiy2016generating} that compares the generater features to those from a pretrained comparator network (typically AlexNet trained on ImageNet). The approach induces strong regularization on the possible signals produced. In other words, GAN-based activation maximization does not start the optimization process from scratch, but from a generator model that has been trained for the same or a similar task \cite{dosovitskiy2016generating}.
More specifically, \cite{nguyen2016synthesizing} trains the generator network on ImageNet and activation maximization in some target (ImageNet) network is achieved by optimizing a high-level code (\ie fc\_6) of this generator network. Activation maximization results produced by GANs offer visually impressive results, because the GAN enforces natural looking images and these methods do not have to use extra regularization terms to suppress extremely high input signals, high frequency patterns or translated copies of similar patterns that highly activate some neuron. However, the produced result of this maximization technique is in direct correspondence to the generator, the data used to train this model, and not a random sample from the network under inspection (which serves as a condition for the learned generative prior). Since we are interested in the raw input that excites our representations, we do not employ any generative priors in this paper. In contrast, our approach directly optimizes the spatiotemporal input of the models starting from randomly initialized noise image (appearance) and video (motion) inputs.

\section{Approach} \label{sec:approach}

\begin{figure*}[t]
	\vspace{-10pt}
	\centering
	\includegraphics[width=.9\linewidth]{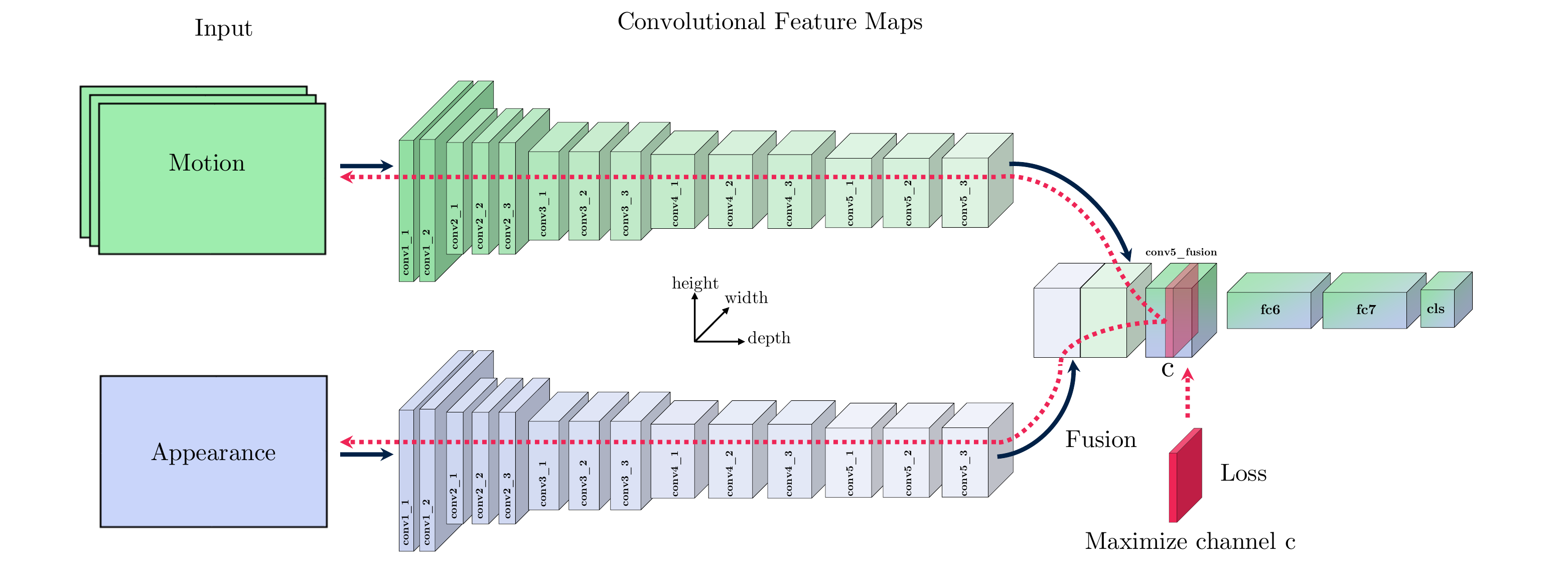}
	\caption[caption]{Schematic of our two-stream activation maximization approach (see Section \protect\ref{sec:approach} for details).}
	\label{fig:vis_architecture}
		\vspace{-10pt}
\end{figure*}

There are several techniques that perform activation maximization for image classification ConvNets \cite{Szegedy14,Simonyan14a,mordvintsev2015inceptionism,yosinski2015understanding,Mahendran16}. We build on these methods and extend them for the spacetime domain to find the preferred spatiotemporal input of individual units in a Two-Stream Fusion model \cite{feichtenhofer2016convolutional}. We formulate the problem as a (regularized) gradient-based optimization problem that searches in the input space. An overview of our approach is shown in \figref{fig:vis_architecture}. A randomly initialized input is presented to the optical flow and the appearance pathways of our model. We compute the feature maps up to a particular layer that  we would like to visualize. 
A single target feature channel, $c$, is selected and activation maximization is performed to generate the preferred input in two steps. First, the derivatives on the input that affect $c$ is calculated by backprogating the target loss, summed over all locations, to the input layer. Second, the propagated gradient is scaled by the learning rate and added to the current input.
These operations are illustrated by the dotted red line in \figref{fig:vis_architecture}. Gradient-based optimization performs these steps iteratively with an adaptively decreasing learning rate until the input converges. Importantly, during this optimization process the network weights are not altered, only the input receives changes. The detailed procedure is outlined in the remainder of this section.

\subsection{Activation maximization}
To make the above more concrete, activation maximization of unit $c$ at layer $l$ seeks an input ${\bx}^* \in\real^{H \times W \times T \times C} $, with $H$ being the height, $W$ the width,  $T$ the duration, and $C$ the color and optical flow channels of the input. We find ${\bx}^*$ by optimizing the following objective  
\begin{equation}\label{eq:vis_objective}
{\bx}^* = \argmax_{\bx }
  \frac{1}{\rho_l^2\ve{\hat{a}}_{l,c} } \langle \ve{a}_l(\bx), e_c  \rangle - \lambda_r \mathcal{R}_r(\bx)
\end{equation}
where $\ve{a}_l$ are the activations at layer $l$, $e_c$ is the natural basis vector corresponding to the $c^\text{th}$ feature channel, and $\mathcal{R}_r$ are regularization term(s) with weight(s) $\lambda_r$. To produce plausible inputs, the unit-specific normalization constant depends on $\rho_l$, which is the size of the receptive field at layer $l$ (\ie the input space), and $\ve{\hat{a}}_{l,c}$, which is the maximum activation of $c$ recorded on a validation set. 

Since the space of possible inputs that satisfy \eqref{eq:vis_objective} is vast, and natural signals only occupy a small manifold of this high-dimensional space, we use regularization to constrain the input in terms of range and smoothness to better fit statistics of natural video signals. Specifically, we apply the following two regularizers, $\mathcal{R}_B$ and $\mathcal{R}_{TV}$, explicitly to the appearance and motion input of our networks. 

\subsection{Regularizing local energy}

As first regularizer, $\mathcal{R}_B$, we enforce a local norm that penalizes large input values

\begin{equation}\label{eq:vis_norm}
\mathcal{R}_B(\bx) = 
 \begin{cases}
 N_B(\bx)
 & \forall i,j,k : \sqrt{\sum_{d} \bx(i,j,k,d)^2} \leq B\\
+\infty, & \text{otherwise}.
\end{cases}
\end{equation}
with 
$
N_B(\bx) = 
\sum_{i,j}\left(\sum_{d} \bx(i,j,k,d)^2\right)^\frac{\alpha}{2}
$
and $i,j,k$ are spatiotemporal indices of the input volume and $d$ indexes either color channels for appearance input, or optical flow channels for motion input, $B$ is the allowed range of the input, and $\alpha$ the exponent of the norm. Similar norms are also used in \cite{Simonyan14a,yosinski2015understanding,Mahendran16}, with the motivation of preventing extreme input scales from dominating the visualization.

\subsection{Regularizing local frequency}

The second regularizer,  $\mathcal{R}_{TV}$,  penalizes high frequency content in the input, since natural signals tend to be dominated by low frequencies. We use a total variation regularizer based on spatiotemporal image gradients 
\begin{align}\label{eq:vis_tv2d3d} 
\mathcal{R}_{TV}(\bx;\kappa,\chi)
=   
\sum_{ijkd} \left[ \kappa  \left( 
(\nabla_x \bx)^2
+
(\nabla_y \bx)^2  \right)
+
\chi 
(\nabla_t \bx)^2  \right],
\end{align}
where $i, j, k$ are used to index the spatiotemporal dimensions of input $\bx$, $d$ indexes the color and optical flow channels of the input, and $\nabla_x, \nabla_y$, $\nabla_t$ are the derivative operators in the horizontal, vertical and temporal direction, respectively. $\kappa$ is used for weighting the degree of spatiotemporal variation and 
 $\chi$ is an explicit slowness parameter that accounts for the regularization strength on the temporal frequency. By varying $0 \le \chi < \infty$ we can selectively penalize with respect to the slowness of the features at the input.

We now derive interesting special cases of \eqref{eq:vis_tv2d3d} that we will investigate in our experiments:
\vspace{-5pt}
\begin{itemize}[noitemsep]

    \item A purely spatial regularizer, $\kappa >0; \chi = 0$ does not penalize variation over the temporal dimension, $t$. This choice produces reconstructions with unconstrained temporal frequency while only enforcing two-dimensional spatial smoothness in \eqref{eq:vis_tv2d3d}.  This choice can be seen as an implicit low-pass filtering in the 2D spatial domain.
    
    \item An isotropic spatiotemporal regularizer, $\kappa = \chi; \kappa,\chi > 0 $ equally penalizes variation in space and time. This can be seen as an implicit low-pass filtering in the 3D spatiotemporal domain.
   
    \item An anisotropic spatiotemporal regularizer, $\kappa \neq \chi; \kappa,\chi > 0  $ allows balancing between space and time to \eg visualize fast varying features in time that are smooth in space. The isotropic case above would bias the visualization to be smooth both in space and time, but not allow us to trade-off between the two.

\end{itemize}

\noindent\textbf{Discussion.}
Purely spatial variation regularization is important to reconstruct natural images, examples of application include image/video restoration \cite{zhang2010non}, feature inversion \cite{Mahendran16}, or style transfer \cite{johnson2016perceptual}, or activation maximization \cite{yosinski2015understanding} where a 2D Gaussian filter was applied after each maximization iteration to achieve a similar effect. 
Isotropic spatiotemporal regularization relates to multiple hand-designed features that operate by derivative filtering of video signals, examples include HOG3D \cite{Klaeser2008}, Cuboids \cite{Dollar05}, or SOEs \cite{Feichtenhofer2015}.
Finally, anisotropic spatiotemporal regularization relates to explicitly modelling the variation in the temporal dimension. Larger weights $\chi$ in \eqref{eq:vis_tv2d3d} stronger penalize the temporal derivative of the signal and consequently enforce low-pass characteristic such that it varies slowly in time. This is a well studied principle in the literature.  For learning general representations from video in an unsupervised manner, minimizing the variation across time is seen both in biological, \eg \cite{foldiak1991learning, wiskott2002slow}, and artificial, \eg, \cite{goroshin2015unsupervised} systems. The motivation for such an approach comes from how the brain solves object recognition by building a stable, slowly varying feature space with respect to time \cite{wiskott2002slow} in order to model temporally contiguous objects for recognition. 

In summary, the regularization of the objective,  \eqref{eq:vis_objective},  combines \eqref{eq:vis_norm} and \eqref{eq:vis_tv2d3d}: 
$
\mathcal{R}_r(\bx) = \mathcal{R}_B(\bx) + \mathcal{R}_{TV}(\bx;\kappa,\chi).
$
 Thus, $\mathcal{R}_r(\bx)$ serves to bias the visualizations to the space of natural images in terms of their magnitudes and spatiotemporal rates of change. Note that the three different special cases of the variational regularizer for the motion input allow us to reconstruct signals that are varying slowly in space, uniformly in spacetime and non-uniformly in spacetime. 

\subsection{Implementation details}
For optimizing the overall objective, \eqref{eq:vis_objective}, we use ADAM that adaptively scales the gradient updates on the input by its inverse square root, while aggregating the gradients in a sliding window over previous iterations. We use the same initializations as in \cite{Mahendran16}.
During optimization, we spatially sift (jitter) \cite{mordvintsev2015inceptionism} the input randomly between 0 and the stride of the optimized layer. 
For all results shown in this paper, we chose the regularization/loss trade-off factors $\lambda_r$ to provide similar weights for the different terms \eqref{eq:vis_norm} - \eqref{eq:vis_tv2d3d}.  
We apply the regularizers separately to the optical flow and appearance input. The regularization terms for the appearance input are chosen to $\lambda_{B,\text{rgb}}=\frac{1}{H W B^\alpha}$ and  $\lambda_{{TV},\text{rgb}}=\frac{1}{H W V^2}$, with $V = B/6.5$, $B=160$ and $\alpha=3$, \ie the default parameters in \cite{Mahendran16}.
The motion input's regularization differs from that of appearance, as follows. In general, the optical flow is assumed to be smoother than appearance input; therefore, the total-variation regularization term of motion inputs has 10 times higher weight than the one for the appearance input. 
In order to visualize different speeds of motion signals, we use different weight terms for the variational regularizers of the motion input.  In particular, to reconstruct different uniformly regularized spatiotemporal inputs we vary  $\kappa$ for penalizing the degree of spatiotemporal variation for reconstructing the motion input (we set $\chi=\kappa$ and only list the values for $\kappa$ in the experiments). For anisotropic spatiotemporal reconstruction, we vary the temporal slowness parameter, $\chi$ and fix $\kappa=1$.  The values in all visualizations are scaled to min-max over the whole sequence for effectively visualizing the full range of motion.

	\section{Experiments}
For sake of space, we focus all our experimental studies on a VGG-16 two-stream fusion model \cite{feichtenhofer2016convolutional} that is illustrated in \figref{fig:vis_architecture} and trained on UCF-101. Our visualization technique, however, is generally applicable to any spatiotemporal architecture. In the supplementary material\footnote{\url{http://feichtenhofer.github.io/action_vis.pdf}}, we visualize various other architectures: Spatiotemporal Residual Networks \cite{feichtenhofer2016residual} using ResNet50 streams, Temporal Segment Networks \cite{WangECCV16} using BN-Inception \cite{ioffe2015batch} or Inception\_v3 \cite{szegedy2015rethinking} streams, trained on multiple datasets: UCF101 \cite{UCF101}, HMDB51\cite{kuehne2011hmdb} and Kinetics \cite{carreira2017quo}. 

We plot the appearance stream input directly by showing an RGB image and the motion input by showing the optical flow as a video that plays on click; the RGB channels of this video consist of the horizontal, vertical and magnitude of the optical flow vectors, respectively. It is our impression that the presented flow visualization is perceptually easier to understand than standard alternatives (\eg HSV encoding); comparison of alternative flow visualization techniques is provided in the supplementary material.

\subsection{Emergence of spatiotemporal features} \label{sec:exp_fusion_features}
We first study the conv5\_fusion layer (\ie the last local layer; see \figref{fig:vis_architecture} for the overall architecture), which takes in features from the appearance and motion streams and learns a local fusion representation for subsequent fully-connected layers with global receptive fields. Therefore, this layer is of particular interest as it is the first point in the network's forward pass where appearance and motion information come together. At conv5\_fusion we see the emergence of both class specific and class agnostic units (\ie general units that form a distributed representation for multiple classes). We illustrate both of these by example in the following. 

\begin{figure}[t]
	\vspace{-10pt}
	\captionsetup[subfigure]{position=bottom}
	\captionsetup[subfloat]{captionskip=0pt}
	\centering 
	\subfloat[{\tiny appearance slow }]{\includegraphics[width=0.12\textwidth]{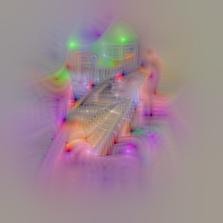}\label{fig_bil_app_slow}}    
	\subfloat[ {\tiny  motion slow}]{\animategraphics[loop,autopause,width=0.12\textwidth]{\theFPS}{figures/conv5_fusion/billiard/intermed1/f021-flow_mag_tv3dx025-}{0}{9}\label{fig_bil_mag_slow}}
	\subfloat[{\tiny appearance fast }]{\includegraphics[width=0.12\textwidth]{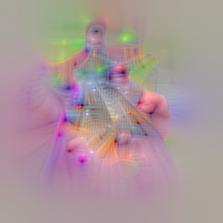}\label{fig_bil_app_fast}}
	\subfloat[ {\tiny  motion fast}]{\animategraphics[loop,autopause,width=0.12\textwidth]{\theFPS}{figures/conv5_fusion/billiard/fast/f021-flow_mag-}{0}{9}\label{fig_bil_mag_fast}}
	
	\caption{Studying the Billiards unit at layer conv5\_fusion from \figref{fig_vis_billiard}. We now show what highly activates the filter in the appearance and in the motion input space using intermediate spatiotemporal variation regularization (a) and (b): Figs.~(c) and (d) show what excites the filter when there is no restriction on the temporal variation of the input: The appearance, (\ref{fig_bil_app_fast}) now also shows a black dot with skin-coloured surroundings at the top which might resemble a head and the motion filter (d) now detects exploding motion patterns (\eg when the white ball hits the others after it has been accelerated by the billiard cue). \color{cyan}{\bf All videos play on click.}
	}
	\label{fig_vis_billiard2}
	\vspace{-10pt}
\end{figure}

\noindent\textbf{Local representation of class specific units.}
In  \figref{fig_vis_billiard} we saw that some local filters might correspond to specific concepts that facilitate recognition of a single class (\eg Billiards). We now reconsider that unit from \figref{fig_vis_billiard} and visualize it under two further spatiotemporal regularization degrees, intermediate and fast temporal variation, in  \figref{fig_vis_billiard2}. (The visualization in \figref{fig_vis_billiard} corresponds to slow motion.) Similar to \figref{fig_vis_billiard}, multiple coloured blobs show up in the appearance, \figref{fig_bil_app_slow}, and  moving circular objects in the motion input (\ref{fig_bil_mag_slow}), but compared to \figref{fig_vis_billiard}, the motion is now varying faster in time. In \figref{fig_bil_mag_fast} and \ref{fig_bil_app_fast}, we only regularize for spatial variation with unconstrained temporal variation, \ie $\chi=0$ in \eqref{eq:vis_tv2d3d}.  We observe that this neuron is fundamentally different in the slow and the fast motion case: It looks for linearly moving circular objects in the slow spatiotemporal variation case, while it looks for an exploding, accelerating motion pattern into various directions in the temporally unconstrained (fast) motion case. It appears that this unit is able to detect a particular spatial pattern of motion, while allowing for a range of speeds and accelerations.  Such an abstraction presumably has value in recognizing an action class with a degree of invariance to exact manner in which it unfolds across time. Another interesting fact is that switching the regularizer for the motion input, also has an impact on the appearance input (\figref{fig_bil_app_slow} \vs \ref{fig_bil_app_fast}) even though the regularization for appearance is held constant. This fact empirically verifies that the fusion unit also expects specific appearance when confronted with particular motion signals.

\begin{figure}[t]
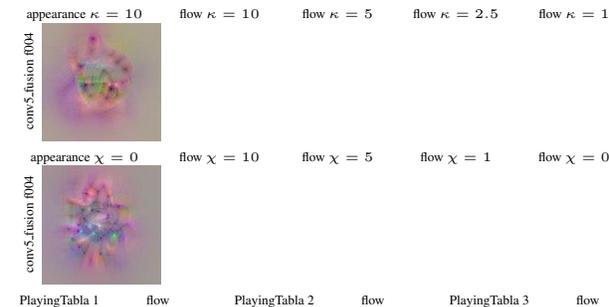

	\vspace{-10pt}
	\captionsetup[subfigure]{labelformat=empty}
	\captionsetup[subfloat]{captionskip=0pt}
	\captionsetup[subfigure]{position=top}
	\centering   
	
	\blockSpeedVisCapUniform{conv5\_fusion}{convfusionSpatial3speeds}{f004}{tv3d}{\kappa}    \vspace{-10pt}
	\blockSpeedVisCap{conv5\_fusion}{convfusionSpatial3speeds}{f004}{tv2d3d}{\chi}
	
	\vspace{-10pt}	
	
	\subfloat[ {\tiny PlayingTabla 1 }]{\animategraphics[loop,autopause,width=0.075\textwidth]{\theFPS}{figures/class/class_flow/c-PlayingTabla_g01_c01/vis_rgb/frame0000}{10}{25}}
	\subfloat[ {\tiny  flow}]{\animategraphics[loop,autopause,width=0.075\textwidth]{\theFPS}{figures/class/class_flow/c-PlayingTabla_g01_c01/vis_mag/frame0000}{10}{25}}
	\hfill
	\subfloat[ {\tiny PlayingTabla 2 }]{\animategraphics[loop,autopause,width=0.075\textwidth]{\theFPS}{figures/class/class_flow/c-PlayingTabla_g02_c01/vis_rgb/frame0000}{10}{25}}
	\subfloat[ {\tiny  flow}]{\animategraphics[loop,autopause,width=0.075\textwidth]{\theFPS}{figures/class/class_flow/c-PlayingTabla_g02_c01/vis_mag/frame0000}{10}{25}}
	\hfill
	\subfloat[ {\tiny PlayingTabla 3 }]{\animategraphics[loop,autopause,width=0.075\textwidth]{\theFPS}{figures/class/class_flow/c-PlayingTabla_g04_c01/vis_rgb/frame0000}{10}{25}}
	\subfloat[ {\tiny flow}]{\animategraphics[loop,autopause,width=0.075\textwidth]{\theFPS}{figures/class/class_flow/c-PlayingTabla_g04_c01/vis_mag/frame0000}{10}{25}}
	
	\caption{Specific unit at conv5\_fusion. Comparison between isotropic and anisotropic spatiotemporal regularization for a single filter at the last convolutional layer. The columns show the appearance and the motion input generated by maximizing the unit, under different degrees of isotropic spatiotemporal ($\kappa$) and anisotropic spatiotemporal TV regularization ($\chi$). The last row shows sample videos of appearance and optical flow  from the PlayingTabla class. }
	\label{fig:vis_playingTabla}
	\vspace{-10pt}
\end{figure}

We now consider unit f004 at conv5\_fusion in \figref{fig:vis_playingTabla}. It seems to capture some drum-like structure in the center of the receptive field, with skin-colored structures in the upper region. This unit could relate to the PlayingTabla class. In \figref{fig:vis_playingTabla} we show the unit under different spacetime regularizers and also show sample frames from three PlayingTabla videos from the test set. Interestingly, when stronger regularization is placed on both spatial and temporal change (\eg $\kappa=10$, top row) we see that a skin colour blob is highlighted in the appearance and a horizontal motion blob is highlighted in the motion in the same area, which combined could capture the characteristic head motions of a drummer. In contrast, with less constraint on motion variation (\eg $\chi=0$, bottom row) we see that the appearance more strongly highlights the drum region, including hand and arm-like structures near and over the drum, while the motion is capturing high frequency oscillation where the hands would strike the drums. Significantly, we see that this single unit fundamentally links appearance and motion: We have the emergence of true spatiotemporal features.

\begin{figure}[t]
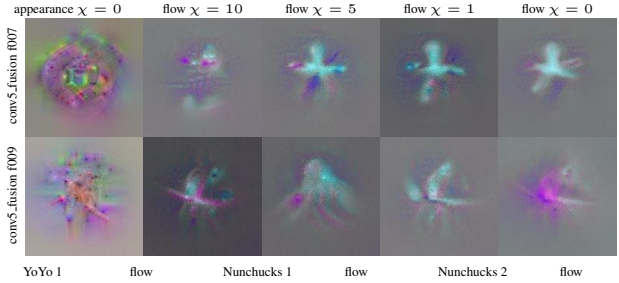

	\vspace{-10pt}
	\captionsetup[subfigure]{labelformat=empty}
	\captionsetup[subfloat]{captionskip=0pt}
	\captionsetup[subfigure]{position=top}
	\centering

	\blockSpeedVisCap{conv5\_fusion}{convfusionSpatial3speeds}{f007}{tv2d3d}{\chi}
	\blockSpeedVisNoCap{conv5\_fusion}{convfusionSpatial3speeds}{f009}{tv2d3d}
	
	\vspace{-10pt}	
	
	\subfloat[ {\tiny YoYo 1 }]{\animategraphics[loop,autopause,width=0.075\textwidth]{\theFPS}{figures/class/class_flow/c-YoYo_g04_c02/vis_rgb/frame0000}{10}{25}}
	\subfloat[ {\tiny  flow}]{\animategraphics[loop,autopause,width=0.075\textwidth]{\theFPS}{figures/class/class_flow/c-YoYo_g04_c02/vis_mag/frame0000}{10}{25}}
	\hfill
	\subfloat[ {\tiny Nunchucks 1 }]{\animategraphics[loop,autopause,width=0.075\textwidth]{\theFPS}{figures/class/class_flow/c-Nunchucks_g03_c01/vis_rgb/frame0000}{10}{25}}
	\subfloat[ {\tiny  flow}]{\animategraphics[loop,autopause,width=0.075\textwidth]{\theFPS}{figures/class/class_flow/c-Nunchucks_g03_c01/vis_mag/frame0000}{10}{25}}
	\hfill
	\subfloat[ {\tiny Nunchucks 2 }]{\animategraphics[loop,autopause,width=0.075\textwidth]{\theFPS}{figures/class/class_flow/c-Nunchucks_g04_c01/vis_rgb/frame0000}{10}{25}}
	\subfloat[ {\tiny  flow}]{\animategraphics[loop,autopause,width=0.075\textwidth]{\theFPS}{figures/class/class_flow/c-Nunchucks_g04_c01/vis_mag/frame0000}{10}{25}}
	
	\caption{Two general units at the convolutional fusion layer. The columns show the appearance and the motion input generated by maximizing the unit, under different degrees of anisotropic spatiotemporal regularization ($\chi$). The last row shows videos of 15 sample frames from the YoYo and Nunchucks classes.}
	\label{fig:vis_general1}
	\vspace{-10pt}
\end{figure}

\begin{figure}[t]
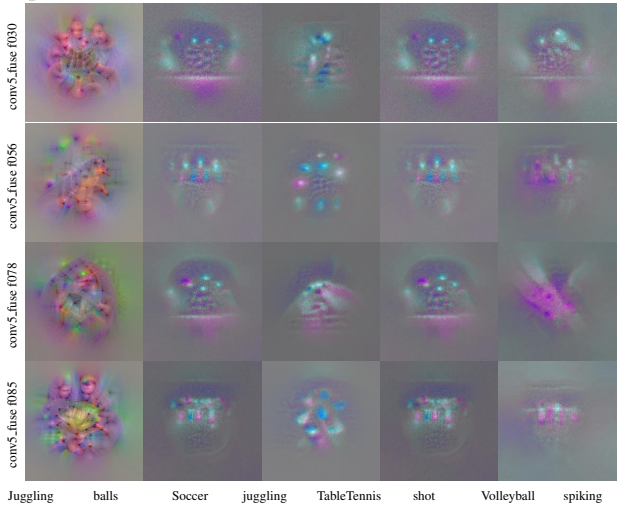

	\captionsetup[subfigure]{labelformat=empty}
	\captionsetup[subfloat]{captionskip=0pt}
	\captionsetup[subfigure]{position=top}
	\centering   
	
	\blockSpeedVisNoCap{conv5\_fuse}{convfusionTemporal3speeds}{f030}{tv2d3d}
	\blockSpeedVisNoCap{conv5\_fuse}{convfusionTemporal3speeds}{f056}{tv2d3d}
	\blockSpeedVisNoCap{conv5\_fuse}{convfusionTemporal3speeds}{f078}{tv2d3d}
	\blockSpeedVisNoCap{conv5\_fuse}{convfusionTemporal3speeds}{f085}{tv2d3d}
	\vspace{-10pt}	
	
	\subfloat[ {\tiny Juggling }]{\animategraphics[loop,autopause,width=0.057\textwidth]{\theFPS}{figures/class/class_flow/c-JugglingBalls_g01_c04/vis_rgb/frame0000}{10}{25}}
	\subfloat[ {\tiny  balls  }]{\animategraphics[loop,autopause,width=0.057\textwidth]{\theFPS}{figures/class/class_flow/c-JugglingBalls_g01_c04/vis_mag/frame0000}{10}{25}}
	\hfill
	\subfloat[ {\tiny Soccer }]{\animategraphics[loop,autopause,width=0.057\textwidth]{\theFPS}{figures/class/class_flow/c-SoccerJuggling_g02_c05/vis_rgb/frame0000}{10}{25}}
	\subfloat[ {\tiny   juggling }]{\animategraphics[loop,autopause,width=0.057\textwidth]{\theFPS}{figures/class/class_flow/c-SoccerJuggling_g02_c05/vis_mag/frame0000}{10}{25}}
	\hfill
	\subfloat[ {\tiny TableTennis }]{\animategraphics[loop,autopause,width=0.057\textwidth]{\theFPS}{figures/class/class_flow/c-TableTennisShot_g02_c04/vis_rgb/frame0000}{10}{25}}
	\subfloat[ {\tiny  shot  }]{\animategraphics[loop,autopause,width=0.057\textwidth]{\theFPS}{figures/class/class_flow/c-TableTennisShot_g02_c04/vis_mag/frame0000}{10}{25}}
	\hfill
	\subfloat[ {\tiny Volleyball  }]{\animategraphics[loop,autopause,width=0.057\textwidth]{\theFPS}{figures/class/class_flow/c-VolleyballSpiking_g01_c01/vis_rgb/frame0000}{10}{25}}
	\subfloat[ {\tiny   spiking}]{\animategraphics[loop,autopause,width=0.057\textwidth]{\theFPS}{figures/class/class_flow/c-VolleyballSpiking_g01_c01/vis_mag/frame0000}{10}{25}}
	
	\caption{General units at the convolutional fusion layer that could be useful for representing ball sports. The columns show the appearance and the motion input generated by maximizing the unit, under different degrees of temporal regularization ($\chi$). The last row shows sample videos from UCF101. }
	\label{fig:vis_general2}
	\vspace{-15pt}	
\end{figure}

\noindent\textbf{Distributed representation of general units.}
In contrast to units that seem very class specific, we also find units that seem well suited for cross-class representation. To begin, we consider  filters f006 and f009 at the conv5\_fusion layer that fuses from the motion into the appearance stream, as shown in \figref{fig:vis_general1}. 
These units seem to capture general spatiotemporal patterns for recognizing classes such as YoYo and Nunchucks, as seen when comparing the unit visualizations to the sample videos from the test set. Next, in \figref{fig:vis_general2}, we similarly show general feature examples for the conv5\_fusion layer that seem to capture general spatiotemporal patterns for recognizing classes corresponding to multiple ball sport actions such as Soccer or TableTennis. These visualizations reveal that at the last convolutional layer the network builds a local representation that can be both distributed over multiple classes and quite specifically tuned to a particular class (\eg \figref{fig:vis_playingTabla} above).

\subsection{Progressive feature abstraction with depth}
\newcounter{cA} 
\newcounter{cB}
\setcounter{cA}{1}
\setcounter{cB}{1}

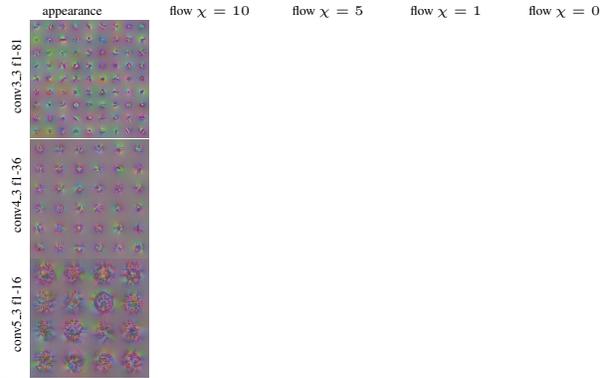
\begin{figure}[t]
	\vspace{-15pt}
	\centering%
	\captionsetup[subfloat]{captionskip=0pt}
	\captionsetup[subfloat]{aboveskip=-20pt}
	\captionsetup[subfloat]{belowskip=-20pt}	
	\captionsetup[subfigure]{labelformat=empty, position=top}
	\subfloat[{\tiny  appearance}]{
		\begin{tikzpicture}%
		\node  {\rotatebox{90}{\tiny \hspace{0.01\textwidth} conv3\_3 f1-81}};
		\end{tikzpicture}%
		\includegraphics[width=0.09\textwidth]{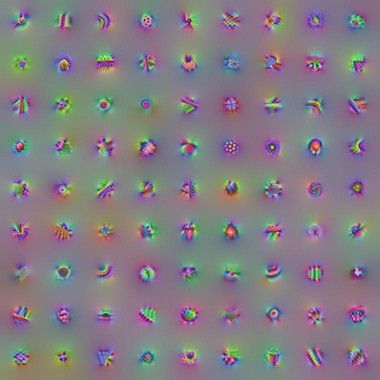}}\hfill
	\subfloat[{\tiny flow  $\chi=10$}]{\animategraphics[loop,autopause,width=0.09\textwidth]{\theFPS}{figures/conv1-5_mosaic_3speeds/conv3_3-mosaic-f1-81-flow_mag_tv2d3d-}{0}{9}}
	\subfloat[{\tiny flow  $\chi=5$}]{\animategraphics[loop,autopause,width=0.09\textwidth]{\theFPS}{figures/conv1-5_mosaic_3speeds/conv3_3-mosaic-f1-81-flow_mag_tv2d3dx05-}{0}{9}}
	\subfloat[{\tiny flow  $\chi=1$}]{\animategraphics[loop,autopause,width=0.09\textwidth]{\theFPS}{figures/conv1-5_mosaic_3speeds/conv3_3-mosaic-f1-81-flow_mag_tv2d3dx01-}{0}{9}}	
	\subfloat[{\tiny flow  $\chi=0$}]{\animategraphics[loop,autopause,width=0.09\textwidth]{\theFPS}{figures/conv1-5_mosaic_3speeds/conv3_3-mosaic-f1-81-flow_mag-}{0}{9}}	
	\vspace{-20pt}
	\subfloat[]{
		\begin{tikzpicture}%
		\node  {\rotatebox{90}{\tiny \hspace{0.01\textwidth} conv4\_3 f1-36}};
		\end{tikzpicture}%
		\includegraphics[width=0.09\textwidth]{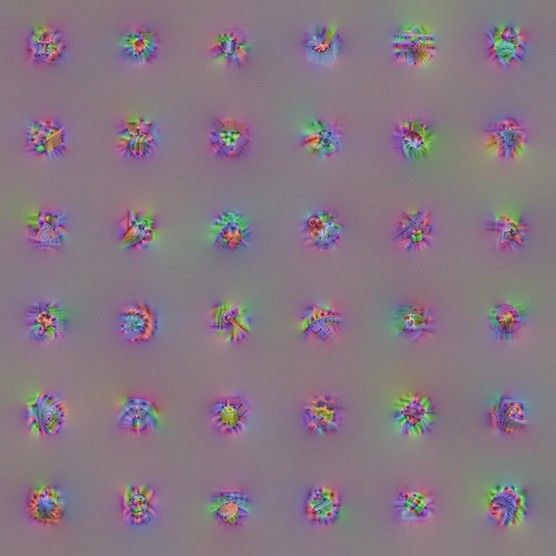}}\hfill
	\subfloat[]{\animategraphics[loop,autopause,width=0.09\textwidth]{\theFPS}{figures/conv1-5_mosaic_3speeds/conv4_3-mosaic-f1-36-flow_mag_tv2d3d-}{0}{9}}
	\subfloat[]{\animategraphics[loop,autopause,width=0.09\textwidth]{\theFPS}{figures/conv1-5_mosaic_3speeds/conv4_3-mosaic-f1-36-flow_mag_tv2d3dx05-}{0}{9}}
	\subfloat[]{\animategraphics[loop,autopause,width=0.09\textwidth]{\theFPS}{figures/conv1-5_mosaic_3speeds/conv4_3-mosaic-f1-36-flow_mag_tv2d3dx01-}{0}{9}}	
	\subfloat[]{\animategraphics[loop,autopause,width=0.09\textwidth]{\theFPS}{figures/conv1-5_mosaic_3speeds/conv4_3-mosaic-f1-36-flow_mag-}{0}{9}}	
	\vspace{-20pt}
	\subfloat[]{
		\begin{tikzpicture}%
		\node  {\rotatebox{90}{\tiny \hspace{0.01\textwidth} conv5\_3 f1-16}};
		\end{tikzpicture}%
		\includegraphics[width=0.09\textwidth]{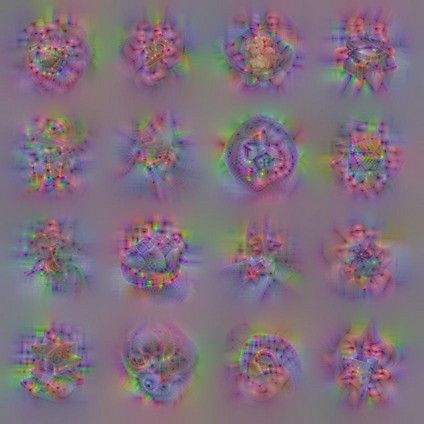}}\hfill
	\subfloat[]{\animategraphics[loop,autopause,width=0.09\textwidth]{\theFPS}{figures/conv1-5_mosaic_3speeds/conv5_3-mosaic-f1-16-flow_mag_tv2d3d-}{0}{9}}
	\subfloat[]{\animategraphics[loop,autopause,width=0.09\textwidth]{\theFPS}{figures/conv1-5_mosaic_3speeds/conv5_3-mosaic-f1-16-flow_mag_tv2d3dx05-}{0}{9}}
	\subfloat[]{\animategraphics[loop,autopause,width=0.09\textwidth]{\theFPS}{figures/conv1-5_mosaic_3speeds/conv5_3-mosaic-f1-16-flow_mag_tv2d3dx01-}{0}{9}}	
	\subfloat[]{\animategraphics[loop,autopause,width=0.09\textwidth]{\theFPS}{figures/conv1-5_mosaic_3speeds/conv5_3-mosaic-f1-16-flow_mag-}{0}{9}}	
	
	\caption{Two-stream conv filters under anisotropic regularization. We show appearance and the optical flow inputs for slowest $\chi=10$, slow $\chi=5$, fast $\chi=1$, and fastest $\chi=0$, temporal variation. Spatial regularization is constant. }
	
	\vspace{-15pt}
	\label{fig:vis_conv1_5_tv2d3d}
\end{figure}

\noindent\textbf{Visualization of early layers.}
We now explore the layers of a VGG-16 Two-Stream architecture \cite{feichtenhofer2016convolutional}. 
In \figref{fig:vis_conv1_5_tv2d3d} we show what excites the convolutional filters of a two-stream architecture at the early layers of the network hierarchy. We use the anisotropic regularization in space and time that penalizes variation at a constant rate across space and varies according to the temporal regularization strength, $\chi$, over time.

We see that the spatial patterns are preserved throughout various temporal regularization factors $\chi$, at all layers. From the temporal perspective, we see that, as expected, for decreasing $\chi$ the temporal variation increases; interestingly, however, the directions of the motion patterns are preserved while the optimal motion magnitude varies with $\chi$. For example, consider the last shown unit f36 of layer conv4\_3 (bottom right filter in the penultimate row of \figref{fig:vis_conv1_5_tv2d3d}). This filter is matched to motion blobs moving in an upward direction.
In the temporally regularized case, $\chi > 0$, the motion is smaller compared to that seen in the temporally unconstrained case, $\chi = 0$. Notably, all these motion patterns strongly excite the same unit. These observations suggest that the network has learned speed invariance, \ie the unit can respond to the same direction of motion with robustness to speed. Such an ability is significant for recognition of actions irrespective of the speed at which they are executed, \eg being able to recognize ``running'' without a concern for how fast the runner moves.
For a comparison of multiple early layer filters under isotropic spatiotemporal regularization please consider the supplementary material.

\noindent\textbf{Visualization of fusion layers.}
We now briefly re-examine the convolutional fusion layer (as in the previous \sref{sec:exp_fusion_features}).
In  \figref{fig:vis_convfusionSpatial3speeds4}, we show the filters at the  conv5\_fusion layer, which fuses from the motion into the appearance stream, while varying the temporal regularization and keeping the spatial regularization constant. This result is again achieved by varying the parameter $\chi$ in \eqref{eq:vis_tv2d3d} visualizations of varying the  regularization strengths isotropically ($\kappa$) are shown in the supplementary material). The visualizations reveal that these first 3 fusion filters at this last convolutional layer show reasonable combinations of appearance and motion information, a qualitative proof that the fusion model in  \cite{feichtenhofer2016convolutional} performs as desired. For example, the receptive field centre of conv5\_fusion f002 seems matched to lip like appearance with a juxtaposed elongated horizontal structure, while the motion is matched to slight up and down motions of the elongation (\eg flute playing).
Once again, we also observe that the units are broadly tuned across temporal input variation (\ie all the different inputs highly activate the same given unit).

\begin{figure}[h]
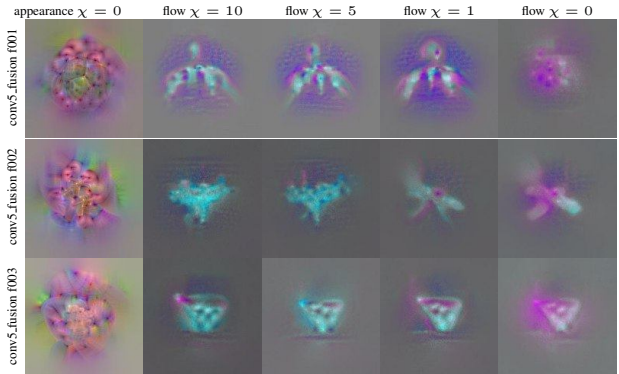
\centering%
	\vspace{-10pt}  
	\captionsetup[subfloat]{captionskip=0pt}
	\captionsetup[subfloat]{aboveskip=-20pt}
	\captionsetup[subfloat]{belowskip=-20pt}	
	\captionsetup[subfigure]{labelformat=empty, position=top}
	
	\foreach \x in {1,...,3}{ 
		\ifnum\thecB<10
		\ifnum\x<2
		\blockSpeedVisCap{conv5\_fusion}{convfusionSpatial3speeds}{f00\thecB}{tv2d3d}{\chi}
		\else
		\blockSpeedVisNoCap{conv5\_fusion}{convfusionSpatial3speeds}{f00\thecB}{tv2d3d}
		\fi
		\else
		\ifnum\x<2
		\blockSpeedVisCap{conv5\_fusion}{convfusionSpatial3speeds}{f0\thecB}{tv2d3d}{\chi}
		\else
		\blockSpeedVisNoCap{conv5\_fusion}{convfusionSpatial3speeds}{f0\thecB}{tv2d3d}
		\fi
		\fi     
		\addtocounter{cB}{1}       
	}
	
	\caption{Visualization of 3 filters of the conv5\_fusion layer. We show the appearance input and the optical flow inputs for slowest $\chi=10$, slow $\chi=5$, fast $\chi=1$, and unconstrained $\chi=0$, temporal variation regularization. The filter in row 2 could be related to the PlayingFlute class by locally filtering lips and a moving instrument (flute). }
	\label{fig:vis_convfusionSpatial3speeds\thecB}
	\vspace{-10pt}
\end{figure}

\noindent\textbf{Visualization of global layers.} We now visualize the layers that have non-local filters, \eg fully-connected layers that operate on top of the convolutional fusion layer illustrated above. \figref{fig:vis_fc6inTemporal3speeds4} and \figref{fig:vis_fc7inTemporal3speeds3} shows filters of the fully-connected layers 6 (fc\_6) and 7 (fc\_7) of the VGG-16 fusion architecture. In contrast to the local features above, we observe a holistic representation that consists of a mixture of the local units seen in the previous layer. For example, in \figref{fig:vis_fc6inTemporal3speeds4} we see units that could combine features for prediction of VolleyballSpiking (top) PlayingFlute (centre) and Archery (bottom row); please compare to the respective prediction layer visualizations in the supplementary material. In \figref{fig:vis_fc7inTemporal3speeds3} we see a unit that resembles the Clean and Jerk action (where a barbell weight is pushed over the head in a standing position) in the top row and another unit that could correspond to Benchpress action (which is performed in lying position on a bench). Notice how the difference in relative body position is captured in the visualizations, \eg the relatively vertical vs. horizontal orientations of the regions captured beneath the weights, especially in the motion visualizations. Here, it is notable that these representations form something akin to a nonlinear (fc\_6) and linear (fc\_7) basis for the prediction layer; therefore, it is plausible that the filters resemble holistic classification patterns. 

\setcounter{cA}{1}
\setcounter{cB}{1}

\begin{figure}[h]
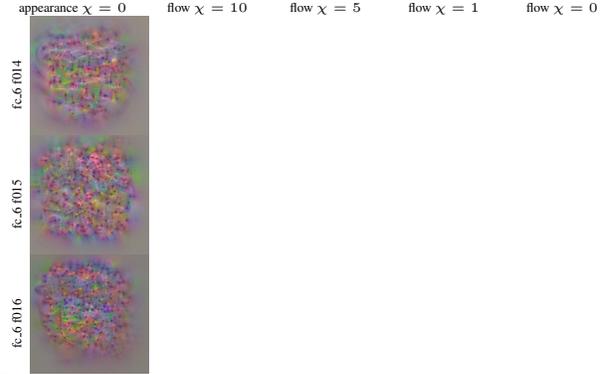
\centering%
	\vspace{-10pt}
	\captionsetup[subfloat]{captionskip=0pt}
	\captionsetup[subfloat]{aboveskip=-20pt}
	\captionsetup[subfloat]{belowskip=-20pt}	
	\captionsetup[subfigure]{labelformat=empty, position=top}
	
	\foreach \x in {14,15,16}{ 
		\ifnum\x<10
		\ifnum\x<2
		\blockSpeedVisCap{\hspace{0.01\textwidth}fc\_6}{fc6inTemporal3speeds}{f00\x}{tv2d3d}{\chi}
		\else
		\blockSpeedVisNoCap{\hspace{0.01\textwidth}fc\_6}{fc6inTemporal3speeds}{f00\x}{tv2d3d}
		\fi
		\else
		\ifnum\x<15
		\blockSpeedVisCap{\hspace{0.01\textwidth}fc\_6}{fc6inTemporal3speeds}{f0\x}{tv2d3d}{\chi}
		\else
		\blockSpeedVisNoCap{\hspace{0.01\textwidth}fc\_6}{fc6inTemporal3speeds}{f0\x}{tv2d3d}
		\fi
		\fi     
		\addtocounter{cB}{1}       
	}
	
	\caption{Visualization of 3 filters of the fc\_6 layer under different temporal regularization. We show the appearance input and the optical flow inputs for slowest $\chi=10$, slow $\chi=5$, fast $\chi=1$, and unconstrained (fastest) $\chi=0$, temporal variation regularization. The filter shown in the first row could resemble the VolleyballSpiking class whereas the filter shown in the second row is visually similar to the unit for predicting PlayingFlute and the last row to the Archery class.}
	\label{fig:vis_fc6inTemporal3speeds\thecB}
\end{figure}

\setcounter{cB}{1}

\begin{figure}[h]
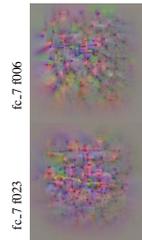
\centering%
	\vspace{-5pt}
	\captionsetup[subfloat]{captionskip=0pt}
	\captionsetup[subfloat]{aboveskip=-20pt}
	\captionsetup[subfloat]{belowskip=-20pt}	
	\captionsetup[subfigure]{labelformat=empty, position=top}
	
	\foreach \x in {6,23}{ 
		\ifnum\x<10
		\ifnum\x<3
		\blockSpeedVisCap{fc\_7}{fc7inTemporal3speeds}{f00\x}{tv2d3d}{\chi}
		\else
		\blockSpeedVisNoCap{fc\_7}{fc7inTemporal3speeds}{f00\x}{tv2d3d}
		\fi
		\else
		\ifnum\x<3
		\blockSpeedVisCap{fc\_7}{fc7inTemporal3speeds}{f0\x}{tv2d3d}{\chi}
		\else
		\blockSpeedVisNoCap{fc\_7}{fc7inTemporal3speeds}{f0\x}{tv2d3d}
		\fi
		\fi     
		\addtocounter{cB}{1}       
	}
	
	\caption{Visualization of 2 filters of the fc\_7 layer under different temporal regularization strength $\chi$. The filter shown in the first row is visually similar to the unit for predicting the Clean and Jerk class in the next layer, whereas the filter shown in the second row could resemble the BenchPress action. }
	\label{fig:vis_fc7inTemporal3speeds\thecB}
	\vspace{-10pt}
\end{figure}

\begin{figure}[h]
	\vspace{-10pt}
	\captionsetup[subfigure]{labelformat=empty}
	\captionsetup[subfloat]{captionskip=0pt}
	\captionsetup[subfloat]{aboveskip=20pt}
	\captionsetup[subfigure]{position=top}
	
	\centering

	\subfloat[{\tiny appearance}]{\includegraphics[width=0.075\textwidth]{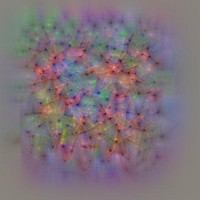}}
	\subfloat[\tiny $\chi=10$]{\animategraphics[loop,autopause,width=0.075\textwidth]{\theFPS}{figures/prediction_3speeds/c-Archery-flow_mag_tv3d-}{0}{9}}
	\subfloat[\tiny $\chi=5$]{\animategraphics[loop,autopause,width=0.075\textwidth]{\theFPS}{figures/prediction_3speeds/c-Archery-flow_mag_tv2d3dx05-}{0}{9}}
	\subfloat[\tiny $\chi=1$]{\animategraphics[loop,autopause,width=0.075\textwidth]{\theFPS}{figures/prediction_3speeds/c-Archery-flow_mag_tv2d3dx01-}{0}{9}}
	\subfloat[\tiny $\chi=0$]{\animategraphics[loop,autopause,width=0.075\textwidth]{\theFPS}{figures/prediction_3speeds/c-Archery-flow_mag-}{0}{9}}
	\hfill
	\subfloat[ {\tiny Archery}]{		\animategraphics[loop,autopause,width=0.08\textwidth]{\theFPS}{figures/class/v_Archery_g01_c01/frame0000}{10}{25}}

	\vspace{-20pt}
	\subfloat[]{\includegraphics[width=0.075\textwidth]{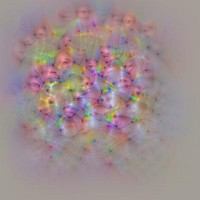}}
	\subfloat[]{\animategraphics[loop,autopause,width=0.075\textwidth]{\theFPS}{figures/prediction_3speeds/c-BabyCrawling-flow_mag_tv3d-}{0}{9}}
	\subfloat[]{\animategraphics[loop,autopause,width=0.075\textwidth]{\theFPS}{figures/prediction_3speeds/c-BabyCrawling-flow_mag_tv2d3dx05-}{0}{9}}
	\subfloat[]{\animategraphics[loop,autopause,width=0.075\textwidth]{\theFPS}{figures/prediction_3speeds/c-BabyCrawling-flow_mag_tv2d3dx01-}{0}{9}}
	\subfloat[]{\animategraphics[loop,autopause,width=0.075\textwidth]{\theFPS}{figures/prediction_3speeds/c-BabyCrawling-flow_mag-}{0}{9}}
	\hfill
	\subfloat[{\tiny BabyCrawling}]{		\animategraphics[loop,autopause,width=0.08\textwidth]{\theFPS}{figures/class/v_BabyCrawling_g01_c01/frame0000}{10}{25}}    
	
	\vspace{-20pt}
	\subfloat[]{\includegraphics[width=0.075\textwidth]{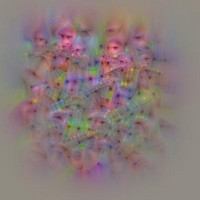}}
	\subfloat[]{\animategraphics[loop,autopause,width=0.075\textwidth]{\theFPS}{figures/prediction_3speeds/c-PlayingFlute-flow_mag_tv3d-}{0}{9}}
	\subfloat[]{\animategraphics[loop,autopause,width=0.075\textwidth]{\theFPS}{figures/prediction_3speeds/c-PlayingFlute-flow_mag_tv2d3dx05-}{0}{9}}
	\subfloat[]{\animategraphics[loop,autopause,width=0.075\textwidth]{\theFPS}{figures/prediction_3speeds/c-PlayingFlute-flow_mag_tv2d3dx01-}{0}{9}}
	\subfloat[]{\animategraphics[loop,autopause,width=0.075\textwidth]{\theFPS}{figures/prediction_3speeds/c-PlayingFlute-flow_mag-}{0}{9}}
	\hfill
	\subfloat[{\tiny PlayingFlute}]{		\animategraphics[loop,autopause,width=0.08\textwidth]{\theFPS}{figures/class/v_PlayingFlute_g01_c01/frame0000}{10}{25}}
	
	\vspace{-20pt}
	\subfloat[]{\includegraphics[width=0.075\textwidth]{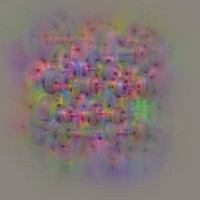}}
	\subfloat[]{\animategraphics[loop,autopause,width=0.075\textwidth]{\theFPS}{figures/prediction_3speeds/c-CleanAndJerk-flow_mag_tv3d-}{0}{9}}
	\subfloat[]{\animategraphics[loop,autopause,width=0.075\textwidth]{\theFPS}{figures/prediction_3speeds/c-CleanAndJerk-flow_mag_tv2d3dx05-}{0}{9}}
	\subfloat[]{\animategraphics[loop,autopause,width=0.075\textwidth]{\theFPS}{figures/prediction_3speeds/c-CleanAndJerk-flow_mag_tv2d3dx01-}{0}{9}}
	\subfloat[]{\animategraphics[loop,autopause,width=0.075\textwidth]{\theFPS}{figures/prediction_3speeds/c-CleanAndJerk-flow_mag-}{0}{9}}
	\hfill
	\subfloat[{\tiny CleanAndJerk}]{		\animategraphics[loop,autopause,width=0.08\textwidth]{\theFPS}{figures/class/v_CleanAndJerk_g01_c01/frame0000}{10}{25}}
	
	\vspace{-20pt}
	\subfloat[]{\includegraphics[width=0.075\textwidth]{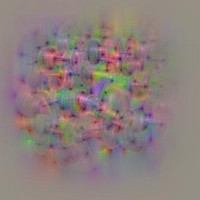}}
	\subfloat[]{\animategraphics[loop,autopause,width=0.075\textwidth]{\theFPS}{figures/prediction_3speeds/c-BenchPress-flow_mag_tv3d-}{0}{9}}
	\subfloat[]{\animategraphics[loop,autopause,width=0.075\textwidth]{\theFPS}{figures/prediction_3speeds/c-BenchPress-flow_mag_tv2d3dx05-}{0}{9}}
	\subfloat[]{\animategraphics[loop,autopause,width=0.075\textwidth]{\theFPS}{figures/prediction_3speeds/c-BenchPress-flow_mag_tv2d3dx01-}{0}{9}}
	\subfloat[]{\animategraphics[loop,autopause,width=0.075\textwidth]{\theFPS}{figures/prediction_3speeds/c-BenchPress-flow_mag-}{0}{9}}
	\hfill
	\subfloat[{\tiny BenchPress}]{		\animategraphics[loop,autopause,width=0.08\textwidth]{\theFPS}{figures/class/v_BenchPress_g01_c01/frame0000}{10}{25}}
	
	\caption{Classification units at the last layer of the network. The first column shows the appearance and the second to fifth columns the motion input generated by maximizing the prediction layer output for the respective classes, with different degrees of temporal variation regularization ($\chi$). The last column shows 15 sample frames from the first video of that class in the test set. }
	\label{fig:vis_class}
\end{figure}

Finally, we visualize the ultimate class prediction layers of the architecture, where the unit outputs corresponds to different classes; thus, we know to what they should be matched. In \figref{fig:vis_class}, we show the fast motion activation of the classes Archery, BabyCrawling, PlayingFlute and CleanAndJerk (see the supplement for additional examples). The learned features for archery (\eg, the elongated bow shape and positioning of the bow as well as the shooting motion of the arrow) are markedly distinct from those of the baby crawling (\eg, capturing the facial parts of the baby appearance while focusing on the arm and head movement in the motion representation), and those of PlayingFlute (\eg filtering eyes and arms (appearance) and moving arms below the flute (motion)), as well as those of CleanAndJerk and BenchPress (\eg capturing barbells and human heads in the appearance with body motion for pressing ($\chi=0$) and balancing ($\chi=10$)  the weight). Thus, we find that the class prediction units have learned representations that are well matched to their classes.

\subsection{Utilizing visualizations for understanding failure modes and dataset bias}
\begin{figure}[h]
	\vspace{-10pt}
	\captionsetup[subfigure]{labelformat=empty}
	\captionsetup[subfloat]{captionskip=0pt}
	\captionsetup[subfloat]{aboveskip=20pt}
	\captionsetup[subfigure]{position=top}
	
	\centering 
	\subfloat[{\tiny appearance}]{\includegraphics[width=0.075\textwidth]{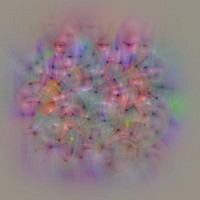}}
	\subfloat[\tiny $\chi=10$]{\animategraphics[loop,autopause,width=0.075\textwidth]{\theFPS}{figures/prediction_3speeds/c-PlayingCello-flow_mag_tv3d-}{0}{9}}
	\subfloat[\tiny $\chi=5$]{\animategraphics[loop,autopause,width=0.075\textwidth]{\theFPS}{figures/prediction_3speeds/c-PlayingCello-flow_mag_tv2d3dx05-}{0}{9}}
	\subfloat[\tiny $\chi=1$]{\animategraphics[loop,autopause,width=0.075\textwidth]{\theFPS}{figures/prediction_3speeds/c-PlayingCello-flow_mag_tv2d3dx01-}{0}{9}}
	\subfloat[\tiny $\chi=0$]{\animategraphics[loop,autopause,width=0.075\textwidth]{\theFPS}{figures/prediction_3speeds/c-PlayingCello-flow_mag-}{0}{9}}
	\hfill
	\subfloat[ {\tiny \mbox{PlayingCello}}]{\animategraphics[loop,autopause,width=0.08\textwidth]{\theFPS}{figures/class/v_PlayingCello_g01_c01/frame0000}{10}{25}}
	
	\vspace{-20pt}
	\subfloat[]{\includegraphics[width=0.075\textwidth]{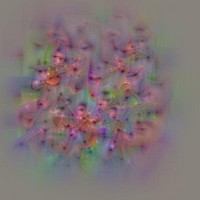}}
	\subfloat[]{\animategraphics[loop,autopause,width=0.075\textwidth]{\theFPS}{figures/prediction_3speeds/c-PlayingViolin-flow_mag_tv3d-}{0}{9}}
	\subfloat[]{\animategraphics[loop,autopause,width=0.075\textwidth]{\theFPS}{figures/prediction_3speeds/c-PlayingViolin-flow_mag_tv2d3dx05-}{0}{9}}
	\subfloat[]{\animategraphics[loop,autopause,width=0.075\textwidth]{\theFPS}{figures/prediction_3speeds/c-PlayingViolin-flow_mag_tv2d3dx01-}{0}{9}}
	\subfloat[]{\animategraphics[loop,autopause,width=0.075\textwidth]{\theFPS}{figures/prediction_3speeds/c-PlayingViolin-flow_mag-}{0}{9}}
	\hfill
	\subfloat[{\tiny PlayingViolin}]{\animategraphics[loop,autopause,width=0.08\textwidth]{\theFPS}{figures/class/v_PlayingViolin_g01_c01/frame0000}{10}{25}}
	
	\caption{Explaining confusion for PlayingCello and PlayingViolin. We see that the learned representation focuses on the horizontal (Cello) and vertical (Violin) alignment of the instrument, which could explain confusions for videos where this is less distinct. }
	\label{fig:vis_cello}
\end{figure}

Another use of our visualizations is to debug the model and reason about failure cases. In UCF101 15\% of the PlayingCello videos get confused as PlayingViolin. In \figref{fig:vis_cello}, we observe that the subtle differences between the classes are related to the alignment of the instruments. In fact, this is in concordance with the confused videos in which the Violins are not aligned in a vertical position. 

\begin{figure}[h]
	\vspace{-10pt}
	\captionsetup[subfigure]{labelformat=empty}
	\captionsetup[subfloat]{captionskip=0pt}
	\captionsetup[subfloat]{aboveskip=20pt}
	\captionsetup[subfigure]{position=top}
	
	\centering 
	\subfloat[{\tiny appearance}]{\includegraphics[width=0.075\textwidth]{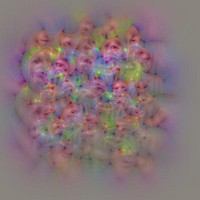}}
	\subfloat[\tiny $\chi=10$]{\animategraphics[loop,autopause,width=0.075\textwidth]{\theFPS}{figures/prediction_3speeds/c-BrushingTeeth-flow_mag_tv3d-}{0}{9}}
	\subfloat[\tiny $\chi=5$]{\animategraphics[loop,autopause,width=0.075\textwidth]{\theFPS}{figures/prediction_3speeds/c-BrushingTeeth-flow_mag_tv2d3dx05-}{0}{9}}
	\subfloat[\tiny $\chi=1$]{\animategraphics[loop,autopause,width=0.075\textwidth]{\theFPS}{figures/prediction_3speeds/c-BrushingTeeth-flow_mag_tv2d3dx01-}{0}{9}}
	\subfloat[\tiny $\chi=0$]{\animategraphics[loop,autopause,width=0.075\textwidth]{\theFPS}{figures/prediction_3speeds/c-BrushingTeeth-flow_mag-}{0}{9}}
	\hfill
	\subfloat[ {\tiny \mbox{BrushingTeeth}}]{\animategraphics[loop,autopause,width=0.08\textwidth]{\theFPS}{figures/class/v_BrushingTeeth_g01_c01/frame0000}{10}{25}}
	
	\vspace{-20pt}
	\subfloat[]{\includegraphics[width=0.075\textwidth]{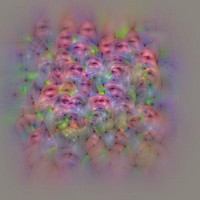}}
	\subfloat[]{\animategraphics[loop,autopause,width=0.075\textwidth]{\theFPS}{figures/prediction_3speeds/c-ShavingBeard-flow_mag_tv3d-}{0}{9}}
	\subfloat[]{\animategraphics[loop,autopause,width=0.075\textwidth]{\theFPS}{figures/prediction_3speeds/c-ShavingBeard-flow_mag_tv2d3dx05-}{0}{9}}
	\subfloat[]{\animategraphics[loop,autopause,width=0.075\textwidth]{\theFPS}{figures/prediction_3speeds/c-ShavingBeard-flow_mag_tv2d3dx01-}{0}{9}}
	\subfloat[]{\animategraphics[loop,autopause,width=0.075\textwidth]{\theFPS}{figures/prediction_3speeds/c-ShavingBeard-flow_mag-}{0}{9}}
	\hfill
	\subfloat[{\tiny ShavingBeard}]{\animategraphics[loop,autopause,width=0.08\textwidth]{\theFPS}{figures/class/v_ShavingBeard_g01_c01/frame0000}{10}{25}}
	
	\caption{Explaining confusion for BrushingTeeth and ShavingBeard. The representation focuses on the local appearance of face and lips as well as the local motion of the tool.  }
	\label{fig:vis_shaving}
	\vspace{-5pt}
\end{figure}

In UCF101 the major confusions are between the classes BrushingTeeth and ShavingBeard. In \figref{fig:vis_shaving} we visualize the inputs that maximally activate these classes and find that they are quite similar, \eg capturing a linear structure moving near the face, but not the minute details that distinguish them. This insight not only explains the confusion, but also can motivate remediation, \eg focused training on the uncaptured critical differences (\ie tooth brush vs shaver).

\begin{figure}[h]
	\vspace{-15pt}
	\captionsetup[subfigure]{labelformat=empty}
	\captionsetup[subfloat]{captionskip=0pt}
	\captionsetup[subfloat]{aboveskip=20pt}
	\captionsetup[subfigure]{position=top}
	
	\centering 
	\subfloat[{\tiny appearance}]{\includegraphics[width=0.075\textwidth]{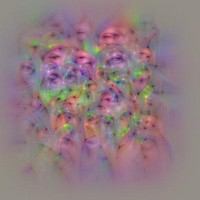}}
	\subfloat[\tiny $\chi=10$]{\animategraphics[loop,autopause,width=0.075\textwidth]{\theFPS}{figures/prediction_3speeds/c-ApplyEyeMakeup-flow_mag_tv3d-}{0}{9}}
	\subfloat[\tiny $\chi=5$]{\animategraphics[loop,autopause,width=0.075\textwidth]{\theFPS}{figures/prediction_3speeds/c-ApplyEyeMakeup-flow_mag_tv2d3dx05-}{0}{9}}
	\subfloat[\tiny $\chi=1$]{\animategraphics[loop,autopause,width=0.075\textwidth]{\theFPS}{figures/prediction_3speeds/c-ApplyEyeMakeup-flow_mag_tv2d3dx01-}{0}{9}}
	\subfloat[\tiny $\chi=0$]{\animategraphics[loop,autopause,width=0.075\textwidth]{\theFPS}{figures/prediction_3speeds/c-ApplyEyeMakeup-flow_mag-}{0}{9}}
	\hfill
	\subfloat[ {\tiny \mbox{ApplyEyeMakeup}}]{\animategraphics[loop,autopause,width=0.08\textwidth]{\theFPS}{figures/class/v_ApplyEyeMakeup_g01_c01/frame0000}{10}{25}}
	
	\vspace{-20pt}
	\subfloat[]{\includegraphics[width=0.075\textwidth]{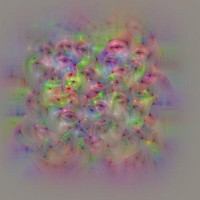}}
	\subfloat[]{\animategraphics[loop,autopause,width=0.075\textwidth]{\theFPS}{figures/prediction_3speeds/c-ApplyLipstick-flow_mag_tv3d-}{0}{9}}
	\subfloat[]{\animategraphics[loop,autopause,width=0.075\textwidth]{\theFPS}{figures/prediction_3speeds/c-ApplyLipstick-flow_mag_tv2d3dx05-}{0}{9}}
	\subfloat[]{\animategraphics[loop,autopause,width=0.075\textwidth]{\theFPS}{figures/prediction_3speeds/c-ApplyLipstick-flow_mag_tv2d3dx01-}{0}{9}}
	\subfloat[]{\animategraphics[loop,autopause,width=0.075\textwidth]{\theFPS}{figures/prediction_3speeds/c-ApplyLipstick-flow_mag-}{0}{9}}
	\hfill
	\subfloat[{\tiny ApplyLipstick}]{\animategraphics[loop,autopause,width=0.08\textwidth]{\theFPS}{figures/class/v_ApplyLipstick_g01_c01/frame0000}{10}{25}}
	
	\caption{Classification units for ApplyEyeMakeup and ApplyLipstick.  Surprisingly, the prediction neuron for ApplyLipstick gets excited by moving eyes at the motion input. Presumably because this resembles a peculiarity of the dataset which contains samples of the ApplyEyeMakeup class with eyes appearing static. }
	\label{fig:vis_apply}
	\vspace{-15pt}
\end{figure}

Dataset bias and generalization to unseen data is important for practical applications. Two classes, ApplyEyeMakeup and ApplyLipstick are, even though being visually very similar, easily classified in the test set of UCF101 with classification rates above 90\% (except for some obvious confusions with BrushingTeeth). This result makes us curious, so we inspect the visualizations in \figref{fig:vis_apply}. The inputs are capturing facial features, such as eyes, and the motion of applicators. Interestingly, it seems that ApplyEyeMakeup and ApplyLipstick are being distinguished, at least in part, by the fact that eyes tend to move in the latter case, while they are held static in the former case. Here, we see a benefit of our visualizations beyond revealing what the network has learned -- they also can reveal idiosyncrasies of the data on which the model has been trained.

	\section{Conclusion}

The compositional structure of deep networks makes it difficult to reason explicitly about what these powerful systems actually have learned. In this paper, we have shed light on the learned representations of deep spatiotemporal networks by visualizing what excites the models internally. We formulate our approach as a  regularized gradient-based optimization problem that searches in the input space of a two-stream architecture by performing activation maximization. We are the first to visualize the hierarchical features learned by a deep motion network. Our visual explanations are highly intuitive and indicate the efficacy of processing appearance and motion in parallel pathways, as well as cross-stream fusion, for analysis of spatiotemporal information. 

\section*{Acknowledgments}
This work was partly supported by  by the Austrian Science Fund (FWF)
under P27076, EPSRC Programme Grant Seebibyte EP/M013774/1, NSERC and CFREF. Christoph \mbox{Feichtenhofer} is a recipient of a DOC Fellowship of the
Austrian Academy of Sciences at the Institute of Electrical
Measurement and Measurement Signal Processing, Graz University of
Technology. 

	{\small
		\bibliographystyle{ieee}
    	\bibliography{bib}

\begin{thebibliography}{10}\itemsep=-1pt

\bibitem{netdissect2017}
D.~Bau, B.~Zhou, A.~Khosla, A.~Oliva, and A.~Torralba.
\newblock Network dissection: Quantifying interpretability of deep visual
  representations.
\newblock In {\em Proc. CVPR}, 2017.

\bibitem{carreira2017quo}
J.~Carreira and A.~Zisserman.
\newblock Quo vadis, action recognition? a new model and the kinetics dataset.
\newblock In {\em Proc. CVPR}, 2017.

\bibitem{Dollar05}
P.~Dollar, V.~Rabaud, G.~Cottrell, and S.~Belongie.
\newblock Behavior recognition via sparse spatio-temporal features.
\newblock In {\em ICCV VS-PETS}, 2005.

\bibitem{dosovitskiy2016generating}
A.~Dosovitskiy and T.~Brox.
\newblock Generating images with perceptual similarity metrics based on deep
  networks.
\newblock In {\em NIPS}, 2016.

\bibitem{Erhan09}
D.~Erhan, Y.~Bengio, A.~Courville, and P.~Vincent.
\newblock Visualizing higher-layer features of a deep network.
\newblock Technical Report 1341, University of Montreal, Jun 2009.

\bibitem{Feichtenhofer2015}
C.~Feichtenhofer, A.~Pinz, and R.~Wildes.
\newblock Dynamically encoded actions based on spacetime saliency.
\newblock In {\em Proc. CVPR}, 2015.

\bibitem{feichtenhofer2016residual}
C.~Feichtenhofer, A.~Pinz, and R.~Wildes.
\newblock Spatiotemporal residual networks for video action recognition.
\newblock In {\em NIPS}, 2016.

\bibitem{feichtenhofer2016convolutional}
C.~Feichtenhofer, A.~Pinz, and A.~Zisserman.
\newblock Convolutional two-stream network fusion for video action recognition.
\newblock In {\em Proc. CVPR}, 2016.

\bibitem{felleman1991distributed}
D.~J. Felleman and D.~C. Van~Essen.
\newblock Distributed hierarchical processing in the primate cerebral cortex.
\newblock {\em Cerebral cortex (New York, NY: 1991)}, 1(1):1--47, 1991.

\bibitem{foldiak1991learning}
P.~F{\"o}ldi{\'a}k.
\newblock Learning invariance from transformation sequences.
\newblock {\em Neural Computation}, 3(2):194--200, 1991.

\bibitem{Goodale92}
M.~A. Goodale and A.~D. Milner.
\newblock Separate visual pathways for perception and action.
\newblock {\em Trends in Neurosciences}, 15(1):20--25, 1992.

\bibitem{goodfellow2014generative}
I.~Goodfellow, J.~Pouget-Abadie, M.~Mirza, B.~Xu, D.~Warde-Farley, S.~Ozair,
  A.~Courville, and Y.~Bengio.
\newblock Generative adversarial nets.
\newblock In {\em NIPS}, 2014.

\bibitem{goroshin2015unsupervised}
R.~Goroshin, J.~Bruna, J.~Tompson, D.~Eigen, and Y.~LeCun.
\newblock Unsupervised feature learning from temporal data.
\newblock In {\em Proc. ICCV}, 2015.

\bibitem{He16}
K.~He, X.~Zhang, S.~Ren, and J.~Sun.
\newblock Deep residual learning for image recognition.
\newblock In {\em Proc. CVPR}, 2016.

\bibitem{Hinton06}
G.~E. Hinton, S.~Osindero, and Y.~W. Teh.
\newblock A fast learning algorithm for deep belief nets.
\newblock {\em Neural Computation}, 18(7):1527--1554, 2006.

\bibitem{ioffe2015batch}
S.~Ioffe and C.~Szegedy.
\newblock Batch normalization: Accelerating deep network training by reducing
  internal covariate shift.
\newblock In {\em Proc. ICML}, 2015.

\bibitem{johnson2016perceptual}
J.~Johnson, A.~Alahi, and L.~Fei-Fei.
\newblock Perceptual losses for real-time style transfer and super-resolution.
\newblock In {\em Proc. ECCV}, 2016.

\bibitem{Klaeser2008}
A.~Kl{\"a}ser, M.~Marsza{\l}ek, and C.~Schmid.
\newblock A spatio-temporal descriptor based on 3d-gradients.
\newblock In {\em Proc. BMVC.}, 2008.

\bibitem{kourtzi2000activation}
Z.~Kourtzi and N.~Kanwisher.
\newblock Activation in human mt/mst by static images with implied motion.
\newblock {\em Journal of cognitive neuroscience}, 12(1):48--55, 2000.

\bibitem{kuehne2011hmdb}
H.~Kuehne, H.~Jhuang, E.~Garrote, T.~Poggio, and T.~Serre.
\newblock {HMDB}: a large video database for human motion recognition.
\newblock In {\em Proc. ICCV}, 2011.

\bibitem{Le12}
Q.~Le, M.~Ranzato, R.~Monga, M.~Devin, K.~Chen, G.~Corrado, J.~Dean, and A.~Ng.
\newblock Building high-level features using large scale unsupervised learning.
\newblock In {\em Proc. ICML}, 2012.

\bibitem{mahendran2016salient}
A.~Mahendran and A.~Vedaldi.
\newblock Salient deconvolutional networks.
\newblock In {\em Proc. ECCV}, 2016.

\bibitem{Mahendran16}
A.~Mahendran and A.~Vedaldi.
\newblock Visualizing deep convolutional neural networks using natural
  pre-images.
\newblock {\em IJCV}, pages 1--23, 2016.

\bibitem{mishkin1983object}
M.~Mishkin, L.~G. Ungerleider, and K.~A. Macko.
\newblock Object vision and spatial vision: two cortical pathways.
\newblock {\em Trends in neurosciences}, 6:414--417, 1983.

\bibitem{mordvintsev2015inceptionism}
A.~Mordvintsev, C.~Olah, and M.~Tyka.
\newblock Inceptionism: Going deeper into neural networks.
\newblock Google Research Blog. Retrieved June 20, 2015.
  \url{https://research.googleblog.com/2015/06/inceptionism-going-deeper-into-neural.html}.

\bibitem{nguyen2016synthesizing}
A.~Nguyen, A.~Dosovitskiy, J.~Yosinski, T.~Brox, and J.~Clune.
\newblock Synthesizing the preferred inputs for neurons in neural networks via
  deep generator networks.
\newblock In {\em NIPS}, 2016.

\bibitem{nguyen2016plug}
A.~Nguyen, J.~Yosinski, Y.~Bengio, A.~Dosovitskiy, and J.~Clune.
\newblock Plug \& play generative networks: Conditional iterative generation of
  images in latent space.
\newblock In {\em Proc. CVPR}, 2017.

\bibitem{nguyen2015deep}
A.~Nguyen, J.~Yosinski, and J.~Clune.
\newblock Deep neural networks are easily fooled: High confidence predictions
  for unrecognizable images.
\newblock In {\em Proc. CVPR}, 2015.

\bibitem{saleem2000connections}
K.~Saleem, W.~Suzuki, K.~Tanaka, and T.~Hashikawa.
\newblock Connections between anterior inferotemporal cortex and superior
  temporal sulcus regions in the macaque monkey.
\newblock {\em Journal of Neuroscience}, 20(13):5083--5101, 2000.

\bibitem{Selvaraju2016}
R.~R. Selvaraju, A.~Das, R.~Vedantam, M.~Cogswell, D.~Parikh, and D.~Batra.
\newblock Grad-cam: Why did you say that? visual explanations from deep
  networks via gradient-based localization.
\newblock {\em arXiv preprint arXiv:1610.02391}, 2016.

\bibitem{Simonyan14a}
K.~Simonyan, A.~Vedaldi, and A.~Zisserman.
\newblock Deep inside convolutional networks: Visualising image classification
  models and saliency maps.
\newblock In {\em ICLR Workshop}, 2014.

\bibitem{Simonyan14b}
K.~Simonyan and A.~Zisserman.
\newblock Two-stream convolutional networks for action recognition in videos.
\newblock In {\em NIPS}, 2014.

\bibitem{Simonyan15}
K.~Simonyan and A.~Zisserman.
\newblock Very deep convolutional networks for large-scale image recognition.
\newblock In {\em International Conference on Learning Representations}, 2015.

\bibitem{UCF101}
K.~Soomro, A.~R. Zamir, and M.~Shah.
\newblock {UCF101}: A dataset of 101 human actions calsses from videos in the
  wild.
\newblock Technical Report CRCV-TR-12-01, 2012.

\bibitem{springenberg2014striving}
J.~T. Springenberg, A.~Dosovitskiy, T.~Brox, and M.~Riedmiller.
\newblock Striving for simplicity: The all convolutional net.
\newblock In {\em ICLR Workshop}, 2015.

\bibitem{szegedy2014going}
C.~Szegedy, W.~Liu, Y.~Jia, P.~Sermanet, S.~Reed, D.~Anguelov, D.~Erhan,
  V.~Vanhoucke, and A.~Rabinovich.
\newblock Going deeper with convolutions.
\newblock In {\em Proc. CVPR}, 2015.

\bibitem{szegedy2015rethinking}
C.~Szegedy, V.~Vanhoucke, S.~Ioffe, J.~Shlens, and Z.~Wojna.
\newblock Rethinking the inception architecture for computer vision.
\newblock {\em arXiv preprint arXiv:1512.00567}, 2015.

\bibitem{Szegedy14}
C.~Szegedy, W.~Zaremba, I.~Sutskever, J.~Bruna, D.~Erhan, I.~Goodfellow, and
  R.~Fergus.
\newblock Intriguing properties of neural networks.
\newblock In {\em Proc. ICLR}, 2014.

\bibitem{C3DICCV2015}
D.~Tran, L.~Bourdev, R.~Fergus, L.~Torresani, and M.~Paluri.
\newblock Learning spatiotemporal features with {3D} convolutional networks.
\newblock In {\em Proc. ICCV}, 2015.

\bibitem{WangECCV16}
L.~Wang, Y.~Xiong, Z.~Wang, Y.~Qiao, D.~Lin, X.~Tang, and L.~{Val Gool}.
\newblock Temporal segment networks: Towards good practices for deep action
  recognition.
\newblock In {\em ECCV}, 2016.

\bibitem{wiskott2002slow}
L.~Wiskott and T.~J. Sejnowski.
\newblock Slow feature analysis: Unsupervised learning of invariances.
\newblock {\em Neural computation}, 14(4):715--770, 2002.

\bibitem{yosinski2015understanding}
J.~Yosinski, J.~Clune, A.~Nguyen, T.~Fuchs, and H.~Lipson.
\newblock Understanding neural networks through deep visualization.
\newblock In {\em ICML Workshop}, 2015.

\bibitem{Zeiler13}
M.~D. Zeiler and R.~Fergus.
\newblock Visualizing and understanding convolutional networks.
\newblock {\em CoRR}, abs/1311.2901, 2013.

\bibitem{zhang2010non}
H.~Zhang, J.~Yang, Y.~Zhang, and T.~S. Huang.
\newblock Non-local kernel regression for image and video restoration.
\newblock In {\em Proc. ECCV}, 2010.

\bibitem{zhang2016top}
J.~Zhang, Z.~Lin, J.~Brandt, X.~Shen, and S.~Sclaroff.
\newblock Top-down neural attention by excitation backprop.
\newblock In {\em Proc. ECCV}, 2016.

\bibitem{zhou2014object}
B.~Zhou, A.~Khosla, A.~Lapedriza, A.~Oliva, and A.~Torralba.
\newblock Object detectors emerge in deep scene cnns.
\newblock In {\em Proc. ICLR}, 2014.

\bibitem{zhou2016learning}
B.~Zhou, A.~Khosla, A.~Lapedriza, A.~Oliva, and A.~Torralba.
\newblock Learning deep features for discriminative localization.
\newblock In {\em Proc. CVPR}, 2016.

\end{thebibliography}
	}

\end{document}